\documentclass{article}

\usepackage{PRIMEarxiv}

\usepackage[utf8]{inputenc} % allow utf-8 input
\usepackage[T1]{fontenc}    % use 8-bit T1 fonts
\usepackage{hyperref}       % hyperlinks
\usepackage{url}            % simple URL typesetting
\usepackage{booktabs}       % professional-quality tables
\usepackage{amsfonts}       % blackboard math symbols
\usepackage{nicefrac}       % compact symbols for 1/2, etc.
\usepackage{microtype}      % microtypography
\usepackage{lipsum}
\usepackage{fancyhdr}       % header
\usepackage{graphicx}       % graphics
\graphicspath{{media/}}     % organize your images and other figures under media/ folder
\usepackage{amsmath}
\usepackage{algorithm, algpseudocode}

\title{Reinforcement Learning for Machine Learning Model Deployment: Evaluating Multi-Armed Bandits in ML Ops Environments
\thanks{\textit{\underline{Preprint Citation}}: McClendon, S.A., Venkatesh, V., Morinelli, J. (2025). Reinforcement Learning for Machine Learning Model Deployment: Evaluating Multi-Armed Bandits in ML Ops Environments. Aimpoint Digital Labs Technical Report.}}

\author{
  S. Aaron McClendon, Vishaal Venkatesh, Juan Morinelli \\
  Aimpoint Digital Labs \\
  \texttt{\{aaron.mcclendon, vishaal.venkatesh, juan.morinelli\}@aimpointdigital.com} 
}

\begin{document}
\maketitle

\begin{abstract}
In modern ML Ops environments, model deployment is a critical process that traditionally relies on static heuristics such as validation error comparisons and A/B testing. However, these methods require human intervention to adapt to real-world deployment challenges, such as model drift or unexpected performance degradation. We investigate whether reinforcement learning, specifically multi-armed bandit (MAB) algorithms, can dynamically manage model deployment decisions more effectively. Our approach enables adaptive production environments by continuously evaluating deployed models and rolling back underperforming ones in real-time. We test six model selection strategies across two real-world datasets and find that RL based approaches match or exceed traditional methods in performance. Our findings suggest that reinforcement learning (RL)-based model management can improve automation, reduce reliance on manual interventions, and mitigate risks associated with post-deployment model failures.
\end{abstract}

% keywords can be removed

\section{Introduction}
Machine learning (ML) models are only as valuable as their ability to perform well in production environments. However, selecting and deploying models in real-world settings presents significant challenges, particularly in \textbf{ML Operations (ML Ops)}. A core issue in ML Ops is \textbf{performance degradation}, often caused by shifts in data distribution over time, commonly referred to as \textbf{drift} \cite{nelson2015modeldrift}.

Drift can manifest in various forms, including \textbf{concept drift} and \textbf{data drift}. 

\textbf{Concept drift} occurs when the statistical properties of the target variable change over time, affecting the relevance of previously learned patterns. A canonical example is spam detection: as adversarial actors adapt their strategies to evade spam filters, models trained on historical spam patterns may gradually lose effectiveness \cite{marchand2025predictivemaintenance}. 

\textbf{Data drift}, on the other hand, refers to shifts in the underlying distribution of input features, leading to discrepancies between training and production data. This phenomenon is well-documented in ML literature \cite{mishra2025mlopsbook}, and it poses a major challenge for models deployed in dynamic environments.

These factors contribute to a critical issue in ML Ops: a model that \textbf{performs well in an offline evaluation setting may fail to generalize effectively once deployed}. Despite substantial research on mitigating model drift in domains such as \textbf{predictive maintenance} \cite{marchand2025predictivemaintenance} and \textbf{healthcare} \cite{xu2025kidneydrift}, many ML Ops workflows still rely on \textbf{static deployment heuristics} that fail to dynamically adjust to real-world conditions.

Currently, model deployment decisions in ML Ops are typically made using \textbf{fixed statistical methods}, such as \textbf{A/B testing} \cite{abtesting,kohavi2017abtesting} or simple validation set error comparisons \cite{mishra2025mlopsbook}. These approaches introduce limitations:
\begin{itemize}
    \item They assume that the distribution of test data \textbf{remains consistent} with the training environment, which is often not the case \cite{xu2025kidneydrift}.
    \item Once a model is deployed, its performance can be \textbf{sparsely reassessed in real time}, often requiring manual intervention and extended delays before performance issues are identified and corrected \cite{hanchuk2025mlopsframework}.
\end{itemize}
These challenges raise fundamental questions: \textbf{When should we deploy a newly trained model into a production environment? How long should we wait until we determine we have a reliable indicator of real world performance?}

Despite these challenges, reinforcement learning-based approaches to model deployment remain largely unexplored in ML Ops research. 

In this work, we investigate whether \textbf{reinforcement learning (RL) \cite{suttonandbarto} methods, specifically multi-armed bandits (MABs) \cite{slivkins2024mab}, can replace traditional static deployment strategies} in ML Ops. Unlike static approaches, MABs dynamically balance exploration (deploying new models) and exploitation (favoring models with strong real-world performance) \cite{slivkins2024mab}, allowing for \textbf{continuous optimization of model deployment} \cite{rauba2025selfhealingml}. Through empirical evaluation on two real-world datasets—\textbf{census wage prediction} and \textbf{fraud detection}—we compare RL-based model deployment to industry-standard methods such as validation error comparisons and A/B testing. Our findings suggest that \textbf{RL-based model selection can offer significant improvements in adaptability, reducing reliance on manual monitoring and mitigating the risks associated with model drift}.

\section{Background}
\label{sec:headings}
Effective model deployment in ML Ops requires a structured approach to evaluating and selecting machine learning models for production. Traditional deployment strategies rely on statistical heuristics such as validation set error comparisons and A/B testing, which assume that a model's offline performance is indicative of its real-world effectiveness. However, these approaches fail to dynamically adjust to changing data distributions and model drift, often leading to performance degradation over time. Reinforcement learning (RL) methods, particularly Multi-Armed Bandits (MABs) \cite{slivkins2024mab}, provide an alternative framework for adaptive model selection by continuously balancing exploration of new models with exploitation of known high-performing models \cite{slivkins2024mab}.

In this study, we compare RL-based \cite{suttonandbarto} model deployment strategies against several commonly used baselines, including naive model replacement, validation error heuristics, and A/B testing. To provide context for the experimental results, this section outlines each of these deployment strategies, explaining their theoretical foundations, industry relevance, and associated advantages and limitations. This discussion serves as a foundation for understanding how reinforcement learning-based approaches differ from traditional ML Ops methodologies and why they may offer improvements in adaptive model selection.

\subsection{Naïve Model Deployment}
One of the simplest strategies for model deployment in ML Ops is the \textbf{naïve method}, in which every newly trained model is automatically deployed to production, replacing the previous version. This approach is frequently used in industry due to its simplicity and automation-friendly workflow, eliminating the need for manual approval or extensive evaluation. Organizations following continuous integration/continuous deployment (CI/CD) practices sometimes adopt this method to ensure models are always using the most recent data and training methodologies.

Despite its ease of implementation, the naïve method carries significant risks. Because it assumes that a newer model will always perform better than its predecessor, it fails to account for \textbf{model drift, overfitting to recent data, or unexpected distribution shifts} in production. Without safeguards, this approach can lead to sudden performance degradation if a newly deployed model generalizes poorly to real-world data. Moreover, organizations using the naïve approach often lack robust rollback mechanisms, making it difficult to recover from unintended performance regressions.

While the naïve method is observed in industry, there is limited academic literature explicitly documenting its usage. However, its prevalence can be inferred from discussions in ML Ops best practices, where automated deployment pipelines are a common feature \cite{mishra2025mlopsbook, hanchuk2025mlopsframework}. This method serves as a baseline for comparison against more adaptive deployment strategies, such as A/B testing and reinforcement learning-based approaches.

\subsection{Validation Testing}
In the validation testing deployment strategy, model selection is based on comparative performance using historical validation scores. Each deployed model is associated with a recorded validation metric, which serves as a benchmark for future model iterations. 

During each retraining cycle, historical data is partitioned into training and validation sets. Let \(\mathcal{D}_t = \{(x_i, y_i)\}_{i=1}^{N_t}\) represent the dataset available at time \(t\), where \(x_i\) are input features and \(y_i\) are target labels. This dataset is split into:
\begin{equation}
    \mathcal{D}_t^{\text{train}}, \mathcal{D}_t^{\text{val}} \sim \mathcal{D}_t, \quad \mathcal{D}_t^{\text{train}} \cap \mathcal{D}_t^{\text{val}} = \emptyset
\end{equation}
where \(\mathcal{D}_t^{\text{train}}\) is used to train the model, and \(\mathcal{D}_t^{\text{val}}\) is used to evaluate performance.

Let \(M_t\) denote the model trained at time \(t\) on \(\mathcal{D}_t^{\text{train}}\). The model's validation performance is computed using an evaluation metric \(f(\cdot)\), such as accuracy, F1-score, or mean squared error:
\begin{equation}
    s_t = f(M_t, \mathcal{D}_t^{\text{val}})
\end{equation}
where \(s_t\) represents the validation score for model \(M_t\).

After a predefined period, a new dataset \(\mathcal{D}_{t+1}\) becomes available, incorporating additional historical data. A new model \(M_{t+1}\) is trained and evaluated:
\begin{equation}
    s_{t+1} = f(M_{t+1}, \mathcal{D}_{t+1}^{\text{val}})
\end{equation}

The validation scores of the new and currently deployed models are then compared:
\begin{equation}
    s_{t+1} > s_t \quad \Rightarrow \quad M_{t+1} \text{ replaces } M_t
\end{equation}
where the new model is deployed only if it demonstrates a superior validation performance.

This method offers a structured and automated approach to model deployment while mitigating the risk of deploying underperforming models. However, it assumes that validation performance is a reliable proxy for real-world effectiveness, an assumption that may not always hold due to factors such as dataset shift and model drift \cite{xu2025kidneydrift}. Additionally, the frequency of retraining and validation testing can influence deployment stability, as models that perform well on static validation sets may still fail to generalize effectively to live data.

\subsection{A/B Testing in ML Ops}
A/B testing is a statistical hypothesis testing framework used to compare two models before committing to a deployment decision in a production environment \cite{abtesting}. The objective is to determine whether a newly trained model, \( M_{t+1} \), provides statistically significant improvements over the currently deployed model, \( M_t \), using new incoming production data.

The currently deployed model \( M_t \) and a newly trained model \( M_{t+1} \) are evaluated on new production data from time step \( t+2 \), denoted as \( \mathcal{D}_{t+2} \). The goal of A/B testing is to test the null hypothesis:

\begin{equation}
    H_0: s_{t+1} \leq s_t
\end{equation}

against the alternative hypothesis:

\begin{equation}
    H_1: s_{t+1} > s_t
\end{equation}

where \( s_t \) and \( s_{t+1} \) represent the performance metrics (e.g., AUC, accuracy, F1-score) of models \( M_t \) and \( M_{t+1} \) respectively on the new production data \( \mathcal{D}_{t+2} \). The null hypothesis \( H_0 \) assumes that the new model does not provide a statistically significant improvement, while \( H_1 \) suggests that the new model is superior.

\subsubsection{Sample Size Estimation and Statistical Power}
To determine whether the difference in performance is significant, we estimate the required sample size, \( n \) needed from \( \mathcal{D}_{t+2} \), based on statistical power analysis. The key parameters involved in A/B testing include:

\begin{itemize}
    \item \textbf{Significance Level} (\(\alpha\)): The probability of a Type I error (false positive), i.e., deploying the new model when it is not actually better.
    \item \textbf{Statistical Power} (\(1 - \beta\)): The probability of correctly rejecting \( H_0 \) when \( H_1 \) is true, where \( \beta \) is the probability of a Type II error (failing to deploy a truly better model).
    \item \textbf{Minimum Detectable Effect} (\(\delta\)): The smallest performance improvement required to consider the new model for deployment.
    \item \textbf{Standard Deviation} (\(\sigma\)): The variability of the model's performance metric.
\end{itemize}

Using the normal approximation, the required sample size \( n \) can be computed using:

\begin{equation}
    n = \left(\frac{(Z_{1-\alpha} + Z_{1-\beta}) \cdot \sigma}{\delta} \right)^2
\end{equation}

where \( Z_{1-\alpha} \) and \( Z_{1-\beta} \) are the critical values from the standard normal distribution corresponding to the significance level and statistical power, respectively.

\subsubsection{A/B Testing Procedure in Deployment}
Once the required sample size \( n \) is determined, A/B testing proceeds as follows:

\begin{algorithm}
    \caption{A/B Testing}
    \begin{algorithmic}[1]
        \Require new production data \( \mathcal{D}_{t+2} \), deployed model \( M_t \), newly trained model \( M_{t+1} \), and hypothesis tolerance \( \delta \)
        \Ensure Deployed model \( M_t \)
        \State Sample \( \mathcal{D}_{t+2} \) (\( n \) samples)
        \State Collect predictions \( \hat{y}_t \) from \( M_t \) and \( \hat{y}_{t+1} \) from \( M_{t+1} \) for these samples
        \State Compute performance metrics \( s_t \) and \( s_{t+1} \) over the \( n \) samples from the new production data.
        \State Perform hypothesis testing:
            \If{\( (s_{t+1} - s_t)/s_t > \delta \) (i.e., if the new model outperforms the current model by at least \( \delta \))}
                \State \( M_t \) $\gets$ \( M_{t+1} \)
            \Else:
                \State \( M_t \) $\gets$ \( M_t \)
            \EndIf
    \end{algorithmic}
\end{algorithm}

This approach ensures that model selection is based on actual production performance rather than validation metrics, which may not accurately reflect real-world conditions.

\subsubsection{Practical Challenges in ML Ops A/B Testing}
In practice, several challenges arise when conducting A/B tests in ML Ops environments:

\begin{itemize}
    \item \textbf{Data Accumulation Delays}: In low-traffic environments, collecting \( n \) data points can take too long, delaying deployment decisions.
    \item \textbf{Statistical Power vs. Timeliness}: Using a large sample size increases statistical confidence but delays adaptation to changes in data distribution.
    \item \textbf{Limited Historical Comparison}: In the adaptation described above, only the current and new models are compared, potentially missing better-performing older models that might be more suitable for the current data distribution. This method can be extended to nAB testing, however this adds a significant amount of complexity which is simplified by the RL methods explored later in this work. 
    \item \textbf{Exploration vs. Exploitation Trade-off}: The focus on statistical significance can limit exploration of promising models that don't immediately show significant improvements.
\end{itemize}

\subsection{Epsilon-Greedy Multi-Armed Bandit}
The \(\epsilon\)-greedy algorithm is one of the simplest and most widely used approaches in the Multi-Armed Bandit (MAB) framework for sequential decision-making \cite{slivkins2024mab}. Unlike traditional statistical methods such as A/B testing \cite{abtesting}, which require large fixed sample sizes to establish significance before deploying a model, \(\epsilon\)-greedy dynamically balances exploration (testing new models) and exploitation (favoring the best-performing model). This makes it particularly well-suited for \textbf{ML Ops environments}, where model performance may shift due to changing data distributions.

\subsubsection{Mathematical Formulation}
At each time step \( t \), the currently deployed model \( M_t \) is selected based on past performance. The model selection process incorporates a probability \( \epsilon \) for exploration (selecting a randomly chosen alternative model) and \( 1 - \epsilon \) for exploitation (choosing the best-performing model so far). Given a set of candidate models \( \mathcal{M} = \{M_1, M_2, ..., M_k\} \), the model \( M_t \) is chosen as:

\begin{equation}
    M_t =
    \begin{cases}
      \arg\max_{M \in \mathcal{M}} f(M, \mathcal{D}_t) & \text{with probability } 1 - \epsilon \\
      M_{\text{random}} \sim \mathcal{M} & \text{with probability } \epsilon
    \end{cases}
\end{equation}

where \( f(M, \mathcal{D}_t) \) represents the evaluation metric (e.g., AUC, accuracy, F1-score) of model \( M \) on dataset \( \mathcal{D}_t \), and \( M_{\text{random}} \) is a randomly selected model from \( \mathcal{M} \).

\subsubsection{Reward Function}
The reinforcement learning component of epsilon-greedy relies on a reward function to evaluate model performance and guide future selections. For each model \( M_i \) at time step \( t \), we maintain an estimated Q-value based on observed performance.

When a model \( M_i \) is selected at batch \( b \) within chunk \( t \) and applied to batch data \( \mathcal{B}_{t,b} \), it receives a binary reward based on performance improvement:

\begin{equation}
    r_i(t,b) = 
    \begin{cases}
        r_{pos}, & \text{if } f(M_i, \mathcal{B}_{t,b}) > f(M_i, \mathcal{B}_{t,b-1}) \text{ or } \\
        r_{neg}, & \text{otherwise}
    \end{cases}
\end{equation}

where \( f(M_i, \mathcal{B}_{t,b}) \) is the performance metric (e.g., AUC-ROC or PR-AUC) of model \( M_i \) on the current batch of data, and \( f(M_i, \mathcal{B}_{t,b-1}) \) is the performance on the previous batch.

For each model, we maintain a running Q-value \( Q_i(t,b) \) that estimates the expected reward. After observing the reward \( r_i(t,b) \), the Q-value is updated using an incremental averaging approach:

\begin{equation}
    Q_i(t,b+1) = Q_i(t,b) + \frac{1}{n_i(t,b)}(r_i(t,b) - Q_i(t,b))
\end{equation}

where \( n_i(t,b) \) is the number of times model \( M_i \) has been selected up to batch \( b \) in chunk \( t \). This update rule gradually adjusts the estimated value toward the true expected reward through repeated observations.

The exploitation term in the epsilon-greedy selection then becomes:

\begin{equation}
    \arg\max_{M_i \in \mathcal{M}_t} Q_i(t,b)
\end{equation}

This selects the model with the highest estimated Q-value based on historical performance. With probability \( \epsilon \), the algorithm instead selects a random model, ensuring continued exploration of the model space.

This binary reward structure incentivizes the selection of models that demonstrate consistent performance improvement across batches. Models that maintain or improve performance receive positive reinforcement, while those that show declining performance receive negative reinforcement, guiding the selection process toward consistently well-performing models.

\subsubsection{Epsilon-Greedy Deployment Process}
The \(\epsilon\)-greedy MAB algorithm proceeds as follows:

\begin{enumerate}
    \item \textbf{Initialize} a set of models \( \mathcal{M} \) with recorded validation scores.
    \item \textbf{For each time step \( t \)}:
    \begin{itemize}
        \item With probability \( 1 - \epsilon \), select the best model \( M_t \) based on highest estimated reward \( \hat{r}_i(t-1) \).
        \item With probability \( \epsilon \), select a random model \( M_{\text{random}} \) to explore potential improvements.
        \item Deploy the selected model on dataset \( \mathcal{D}_t \) and collect real-world performance data.
        \item Calculate reward \( r_i(t) \) and update the model's reward estimate \( \hat{r}_i(t) \) based on observed performance.
    \end{itemize}
    \item Repeat until sufficient evidence is gathered for model selection.
\end{enumerate}

\subsubsection{Comparison to Traditional Statistical Methods}
Unlike \textbf{A/B testing}, which requires a predefined sample size before making a deployment decision \cite{abtesting}, \(\epsilon\)-greedy operates in an \textbf{online learning} setting. This means that:
\begin{itemize}
    \item Decisions are made incrementally rather than waiting for an entire batch of data.
    \item The algorithm continuously adapts as new data arrives, allowing for real-time optimization.
    \item No fixed statistical significance threshold is required; instead, model selection is based on empirical performance over time.
\end{itemize}
While \textbf{validation-based selection} relies purely on offline performance metrics, \(\epsilon\)-greedy incorporates real-world feedback, improving robustness to \textbf{concept drift and dataset shift} \cite{xu2025kidneydrift}.

\subsubsection{Hyperparameters and Practical Considerations}
The key hyperparameter in \(\epsilon\)-greedy is the \textbf{exploration rate} \( \epsilon \), which determines the balance between trying new models and exploiting the best-known model. Several common strategies exist for tuning \( \epsilon \):

\begin{itemize}
    \item \textbf{Fixed \(\epsilon\)}: A constant exploration rate (e.g., \( \epsilon = 0.1 \)), meaning the model explores 10\% of the time and exploits 90\% of the time.
    \item \textbf{Decay \(\epsilon\) Over Time}: Start with a high exploration rate (e.g., \( \epsilon_0 = 1.0 \)) and gradually decrease it over time:
    \begin{equation}
        \epsilon_t = \frac{\epsilon_0}{1 + \lambda t}
    \end{equation}
    where \( \lambda \) controls the rate of decay.
    \item \textbf{Adaptive \(\epsilon\)}: Adjust \( \epsilon \) dynamically based on the confidence interval of model performance.
\end{itemize}

\subsubsection{Challenges in ML Ops Deployment}
Despite its advantages, \(\epsilon\)-greedy presents several challenges in \textbf{ML Ops environments}:

\begin{itemize}
    \item \textbf{Cold Start Problem}: If the initial set of models is suboptimal, early exploitation may reinforce poor decisions.
    \item \textbf{Exploration Cost}: Deploying exploratory models in a live environment introduces risk, as poor-performing models may degrade user experience.
    \item \textbf{Performance Volatility}: Choosing models randomly for exploration can introduce variance in production performance.
\end{itemize}

Despite these challenges, \(\epsilon\)-greedy can provide a lightweight and efficient alternative to static model deployment strategies. By dynamically adjusting to real-world feedback, it enables continuous improvement in \textbf{ML Ops environments}.

\subsection{Upper Confidence Bound (UCB)}
The \textbf{Upper Confidence Bound (UCB)} algorithm is a popular approach in the Multi-Armed Bandit (MAB) framework, designed to balance exploration and exploitation by using confidence intervals to guide model selection \cite{ucb}. Unlike \(\epsilon\)-greedy, which selects models randomly for exploration, UCB prioritizes models with greater uncertainty in their estimated performance, ensuring that underexplored models are evaluated systematically.

\subsubsection{Mathematical Formulation}
At each time step \( t \), we have a set of candidate models \( \mathcal{M} = \{M_1, M_2, ..., M_k\} \), each with an associated evaluation metric \( f(M, \mathcal{D}_t) \) based on dataset \( \mathcal{D}_t \). UCB selects the model \( M_t \) that maximizes the following upper confidence bound:

\begin{equation}
    M_t = \arg\max_{M \in \mathcal{M}} \left( \hat{s}_M + c \sqrt{\frac{\ln t}{N_M}} \right)
\end{equation}

where:
\begin{itemize}
    \item \( \hat{s}_M \) is the estimated performance metric (e.g., AUC, accuracy) of model \( M \) based on previous evaluations.
    \item \( N_M \) is the number of times model \( M \) has been selected up to time \( t \).
    \item \( c \) is a tunable exploration parameter that controls the width of the confidence bound.
    \item \( \ln t \) ensures that models with fewer selections are explored more aggressively as time progresses.
\end{itemize}

The term \( c \sqrt{\frac{\ln t}{N_M}} \) represents the \textbf{exploration bonus}, which encourages selecting models that have been evaluated less frequently, preventing premature convergence to suboptimal models.

\subsubsection{UCB Deployment Process}
The UCB algorithm proceeds as follows:

\begin{enumerate}
    \item \textbf{Initialize} a set of models \( \mathcal{M} \), with no prior evaluations.
    \item \textbf{For each time step \( t \)}:
    \begin{itemize}
        \item Compute the UCB score for each model using Equation (4).
        \item Select the model \( M_t \) that maximizes the UCB score.
        \item Deploy \( M_t \) on dataset \( \mathcal{D}_t \) and collect real-world performance data.
        \item Update \( \hat{s}_{M_t} \) and \( N_{M_t} \) based on observed performance.
    \end{itemize}
    \item Repeat the process until sufficient data has been collected.
\end{enumerate}

\subsubsection{Comparison to Other Methods}
Compared to \textbf{A/B testing} and \textbf{\(\epsilon\)-greedy}, UCB offers a structured approach to exploration, ensuring that underexplored models are prioritized based on confidence intervals rather than random selection \cite{ucb}. The key differences include:

\begin{itemize}
    \item \textbf{Unlike A/B testing}, UCB does not require fixed sample sizes before making deployment decisions; it continuously updates confidence intervals over time.
    \item \textbf{Unlike \(\epsilon\)-greedy}, UCB systematically selects models based on estimated performance uncertainty rather than introducing random exploration.
    \item UCB reduces the risk of over-exploring poor models compared to \(\epsilon\)-greedy, leading to more stable deployments in ML Ops environments.
\end{itemize}

\subsubsection{Hyperparameters and Practical Considerations}
The key hyperparameter in UCB is the \textbf{exploration coefficient} \( c \), which determines the trade-off between exploration and exploitation \cite{ucb}:

\begin{itemize}
    \item \textbf{Higher \( c \)} values encourage more exploration, ensuring that models with high uncertainty are evaluated more frequently.
    \item \textbf{Lower \( c \)} values favor exploitation, quickly converging to the highest-performing model based on current estimates.
\end{itemize}

In practical deployments, choosing \( c \) appropriately is crucial. If \( c \) is too large, the algorithm may spend too much time evaluating weak models. If \( c \) is too small, the system may converge prematurely to a suboptimal model.

\subsubsection{Challenges in ML Ops Deployment}
While UCB provides a systematic approach to model selection, it presents challenges in an ML Ops setting:

\begin{itemize}
    \item \textbf{Cold Start Problem}: Similar to \(\epsilon\)-greedy, UCB requires an initial phase where all models are explored at least once before meaningful confidence bounds can be established.
    \item \textbf{Computational Overhead}: The need to maintain confidence intervals and update model statistics at each step can introduce additional computation costs.
    \item \textbf{Sensitivity to Exploration Parameter}: Choosing an inappropriate value for \( c \) can lead to excessive exploration or premature convergence.
\end{itemize}

\subsection{Thompson Sampling}
Thompson Sampling (TS) is a Bayesian approach to the Multi-Armed Bandit (MAB) problem, which selects models probabilistically based on their estimated performance distributions. Unlike \(\epsilon\)-greedy and Upper Confidence Bound (UCB), which rely on fixed exploration-exploitation trade-offs, Thompson Sampling maintains a posterior distribution over each model's reward and samples from these distributions to guide selection \cite{thompson}. This enables more efficient and adaptive exploration in ML Ops environments.

\subsubsection{Mathematical Formulation}
At time step \( t \), let \( \mathcal{M} = \{M_1, M_2, ..., M_k\} \) represent the set of candidate models. Each model \( M_i \) has an unknown performance metric \( s_i \), which is treated as a random variable with a prior distribution \( P(s_i) \). The objective is to maintain a posterior belief about \( s_i \) based on observed data.

For each model \( M_i \), we assume a likelihood function \( P(\mathcal{D}_t | s_i) \) given dataset \( \mathcal{D}_t \). Using Bayes' theorem, the posterior is updated as:

\begin{equation}
    P(s_i | \mathcal{D}_t) \propto P(\mathcal{D}_t | s_i) P(s_i)
\end{equation}

At each time step \( t \), Thompson Sampling selects a model by drawing a sample from each model’s posterior distribution:

\begin{equation}
    M_t = \arg\max_{M_i \in \mathcal{M}} \tilde{s}_i, \quad \tilde{s}_i \sim P(s_i | \mathcal{D}_t)
\end{equation}

where \( \tilde{s}_i \) is a randomly sampled performance score for model \( M_i \) from its posterior distribution.

\subsubsection{Thompson Sampling Deployment Process}
The Thompson Sampling algorithm proceeds as follows:

\begin{enumerate}
    \item \textbf{Initialize} prior distributions \( P(s_i) \) for each model \( M_i \).
    \item \textbf{For each time step \( t \)}:
    \begin{itemize}
        \item Sample \( \tilde{s}_i \) from the posterior \( P(s_i | \mathcal{D}_t) \) for each model.
        \item Select the model \( M_t \) that maximizes \( \tilde{s}_i \).
        \item Deploy \( M_t \) on dataset \( \mathcal{D}_t \) and observe real-world performance.
        \item Update the posterior \( P(s_i | \mathcal{D}_t) \) using Bayes’ rule.
    \end{itemize}
    \item Repeat until convergence or deployment criteria are met.
\end{enumerate}

\subsubsection{Comparison to Other Methods}
Compared to \textbf{A/B testing}, \(\epsilon\)-greedy, and UCB, Thompson Sampling offers a more \textbf{probabilistic approach} to balancing exploration and exploitation \cite{thompson}:

\begin{itemize}
    \item \textbf{Unlike A/B testing}, Thompson Sampling does not require predefined sample sizes and continuously adapts as new data is collected.
    \item \textbf{Unlike \(\epsilon\)-greedy}, Thompson Sampling does not rely on fixed probabilities for exploration but instead adjusts based on model uncertainty.
    \item \textbf{Unlike UCB}, which selects models based on upper confidence bounds, Thompson Sampling selects models in proportion to their probability of being optimal, leading to more efficient exploration.
\end{itemize}

\subsubsection{Hyperparameters and Practical Considerations}
The performance of Thompson Sampling depends on the choice of prior distributions and update rules \cite{thompson}:

\begin{itemize}
    \item \textbf{Beta-Bernoulli Model}: For binary classification problems, model performance can be treated as a Bernoulli variable with a Beta prior:
    \begin{equation}
        P(s_i) = \text{Beta}(\alpha_i, \beta_i)
    \end{equation}
    where \(\alpha_i\) and \(\beta_i\) are updated based on observed successes and failures.
    
    \item \textbf{Gaussian-Normal Model}: For continuous metrics such as AUC or accuracy, a Gaussian prior with known variance can be used:
    \begin{equation}
        P(s_i) = \mathcal{N}(\mu_i, \sigma_i^2)
    \end{equation}
    where \( \mu_i \) is updated based on the sample mean of observed data.

    \item \textbf{Dirichlet-Multinomial Model}: Used for multi-class classification problems where categorical outcomes influence model performance.

    \item \textbf{Bounded Continuous Metrics}: For performance metrics bounded between [0,1] such as AUC, PRAUC, or accuracy, several modeling approaches exist:
    \begin{itemize}
        \item \textbf{Beta Distribution}: Theoretically, a Beta distribution may be more appropriate for modeling metrics constrained to [0,1]:
        \begin{equation}
            P(s_i) = \text{Beta}(a_i, b_i)
        \end{equation}
        where parameters $a_i$ and $b_i$ are updated based on observed performance.
        
        \item \textbf{Truncated Gaussian}: A truncated normal distribution can ensure samples remain within valid bounds.
        
        \item \textbf{Gaussian-Gamma Model}: In our implementation, we adopt a Gaussian-Gamma conjugate model for the posterior sampling step. While the actual observed PRAUC/AUC values are always bounded between [0,1], the posterior sampling process could theoretically generate values outside this range. Our decision to use this model is justified by several practical advantages:
        \begin{itemize}
            \item Efficient exploration in early stages when uncertainty is high
            \item Posterior distributions that rapidly concentrate within valid bounds as observations accumulate
            \item Adaptive control of exploration via the precision parameter
            \item Computational efficiency and established convergence properties
        \end{itemize}
    \end{itemize}
    Our empirical results demonstrate that the Gaussian-Gamma approach performs well for bounded metrics despite its theoretical limitations. The update rule also naturally penalizes models with unstable performance, providing implicit regularization toward reliable model selection.
\end{itemize}

\subsubsection{Challenges in ML Ops Deployment}
While Thompson Sampling provides a principled approach to model selection, it introduces several challenges in an ML Ops setting:

\begin{itemize}
    \item \textbf{Computational Complexity}: Maintaining and updating posterior distributions for multiple models can be computationally expensive.
    \item \textbf{Cold Start Problem}: Poor initial priors may result in suboptimal early selections, delaying convergence to the best-performing model.
    \item \textbf{Stochastic Model Selection}: Unlike UCB, which always selects the model with the highest confidence bound, Thompson Sampling introduces randomness, which may lead to temporary performance fluctuations.
\end{itemize}

\section{Related Work}
The field of \textbf{ML Operations (ML Ops)} has received increasing attention in recent years as organizations seek to automate and optimize the deployment of machine learning models in production. While the ML Ops process itself is well-studied, the specific challenge of using \textbf{reinforcement learning (RL)} for dynamic model deployment remains largely unexplored. In this section, we review prior work on ML Ops frameworks, model drift, A/B testing, and reinforcement learning-based decision-making, highlighting the gaps our research aims to address.

\subsection{ML Ops and Model Deployment}
Several studies have examined the principles and best practices of ML Ops, focusing on the lifecycle management of production machine learning systems. Mishra \cite{mishra2025mlopsbook} provides a comprehensive guide to implementing ML Ops pipelines, emphasizing automation, monitoring, and retraining strategies. Similarly, Hanchuk and Semerikov \cite{hanchuk2025mlopsframework} present a meta-synthesis of ML Ops tools and frameworks, analyzing common design patterns and deployment architectures.

Despite these contributions, existing research has primarily focused on \textbf{static deployment strategies}, where models are selected and deployed using pre-defined heuristics. Our study differs in that we explore an \textbf{adaptive model selection framework} using RL, allowing the system to dynamically choose the best-performing model in real-time.

\subsection{Model Drift and Performance Degradation}
One of the key challenges in ML Ops is \textbf{model drift}, where changes in data distributions degrade model performance over time. Nelson et al. \cite{nelson2015modeldrift} investigate how various machine learning algorithms respond to adversarial data drift, demonstrating the risks associated with static deployment policies. Similarly, Xu et al. \cite{xu2025kidneydrift} examine mitigation strategies for model drift in healthcare applications, proposing periodic retraining and monitoring techniques.

\subsection{A/B Testing and Traditional Model Selection}
A/B testing is a widely used statistical technique for model selection in production environments. Kohavi and Longbotham \cite{kohavi2017abtesting} provide an overview of controlled online experiments, emphasizing the importance of statistical power and sample size estimation in deployment decisions.

While A/B testing provides a rigorous framework for evaluating new models, it suffers from \textbf{long evaluation times} and the risk of exposing users to underperforming models. In contrast, our approach \textbf{leverages RL methods such as Multi-Armed Bandits (MABs)} to continuously optimize model selection in an online learning setting, reducing reliance on fixed sample sizes and significance thresholds.

\subsection{Multi-Armed Bandits and Reinforcement Learning in ML Ops}
The Multi-Armed Bandit (MAB) framework has been extensively studied in the context of sequential decision-making \cite{slivkins2024mab,contextualmab}. Slivkins \cite{slivkins2024mab} provides a detailed introduction to MABs, covering both stochastic and adversarial settings. While MAB algorithms have been widely applied in areas such as advertising, recommendation systems, and clinical trials, \cite{contextualmab} their application in ML Ops and model deployment remains largely unexplored.

Rauba et al. \cite{rauba2025selfhealingml} introduce the concept of \textbf{self-healing machine learning}, where systems autonomously adapt to changing environments. While their work focuses on model adaptation, it does not address the problem of \textbf{selecting and deploying models in an ML Ops pipeline using RL techniques}. Our work builds on these ideas by investigating how RL-based agents can actively manage the production lifecycle of ML models.

\subsection{Novelty of Our Work}
To the best of our knowledge, there has been no rigorous, quantitative comparison of \textbf{validation-based model selection, A/B testing, and Multi-Armed Bandit methods in ML Ops settings}. Moreover, existing research has not explored whether \textbf{reinforcement learning agents can effectively manage production model deployment}. Our study aims to fill this gap by conducting a systematic evaluation of naïve deployment, validation-based heuristics, A/B testing, and reinforcement learning-based methods (Epsilon-Greedy, UCB, and Thompson Sampling). By comparing these approaches across real-world datasets, we provide new insights into the feasibility of using RL for adaptive model selection in ML Ops.

\section{Experimental Setup and Results}
To evaluate the effectiveness of reinforcement learning-based model deployment strategies in ML Ops, we conducted experiments using two large-scale publicly available datasets: the \textbf{Census Bureau USA dataset} \cite{census2023} and the \textbf{Bank Account Fraud dataset} from NeurIPS 2022 \cite{fraud2022}. These datasets were chosen due to their size and suitability for simulating real-world ML Ops environments. The underlying models were trained using standard ML practitioner methods, such as one-hot encoding of categorical variables, XGBoost classifiers, SMOTE, and some basic feature engineering. Model training pipelines were taken from high performing Kaggle kernels to ensure a fair evaluation of the underlying models. Importantly, the purpose of this experiment was to evaluate the performance of model management methodology, rather than optimizing the underlying ML model training process itself. 

\subsection{Datasets}
\textbf{Census Bureau USA Dataset} \cite{census2023}: This dataset contains demographic and economic information from the U.S. Census Bureau. It is commonly used for predictive modeling tasks such as income classification and labor force participation prediction.In this paper we investigate the classification task of predicting wage in a bucketed binary task, under or over \$50,000 USD per year. This is a slightly imbalanced task, with those earning under \$50,000 USD per year constituting $\approx76\%$ of the dataset.

\textbf{Bank Account Fraud Dataset} \cite{fraud2022}: This dataset consists of anonymized transaction records labeled as fraudulent or legitimate. It was originally published for the NeurIPS 2022 competition on fraud detection, making it suitable for evaluating model drift in financial applications. This dataset is heavily imbalanced, with only 1.1\% of cases being detected as fraudulent. 

\subsection{Simulating an ML Ops Environment}
To test the various ML Ops deployment strategies outlined in the Background section, we simulated a dynamic production environment using sequential data partitions. Each dataset was divided into \textbf{8 sequential chunks of fixed length}, ensuring that models encountered a continuous stream of new data over time. 

The decision to use 8 chunks was made to ensure that each chunk contained a sufficiently large number of data points, allowing newly trained models to have enough historical data to learn meaningful patterns. Additionally, larger chunks provided sufficient intra-chunk time for ML Ops methods to dynamically select models, preventing overly frequent swaps and allowing for more stable performance assessments. Both datasets were divided into 8 chunks to enable direct comparisons between model deployment strategies across different domains.

\subsubsection{Dual Time Scales: Chunk-Level and Batch-Level}
The experimental setup operates across two distinct time scales:
\begin{itemize}
    \item \textbf{Chunk-Level Time Scale (\( t \))}: At each chunk boundary \( t \), a new model \( M_t \) is trained using all available historical data up to and including chunk \( t \):
    \begin{equation}
        \mathcal{D}_t = \bigcup_{i=0}^{t} \mathcal{D}_i
    \end{equation}
    This ensures that newer models are trained on progressively larger datasets, mimicking an ML Ops retraining cycle.
    
    \item \textbf{Batch-Level Time Scale (\( b \))}: Within each chunk \( t \), data arrives in sequential batches, and ML Ops methods dynamically select a model from the available set \( \mathcal{M}_t = \{M_0, M_1, ..., M_t\} \) to process each incoming batch.
    Formally, for each batch \( \mathcal{B}_b \) in chunk \( t \):
    \begin{equation}
        M_{t,b} = \text{SelectModel}(\mathcal{M}_t, \mathcal{B}_b)
    \end{equation}
    where \( \text{SelectModel}(\cdot) \) represents the model selection rule determined by the chosen deployment strategy (e.g., A/B testing, MAB algorithms).
\end{itemize}

This distinction between chunk-level and batch-level time scales is important for understanding how our reinforcement learning algorithms operate. All model selection algorithms (UCB, Thompson Sampling, and $\epsilon$-greedy) make decisions at the batch level but use information accumulated across both time scales. Specifically, the reward functions and model performance estimates discussed in previous sections refer to the batch-level selections, where \( b \) is the time step used in those formulations.

% \subsubsection{Dual Time Scales: Chunk-Level and Intra-Chunk Level}
% The experimental setup operates across two distinct time scales:

% \begin{itemize}
%     \item \textbf{Chunk-Level Time Scale (\( t \))}: At each chunk boundary \( t \), a new model \( M_t \) is trained using all available historical data up to and including chunk \( t \):

%     \begin{equation}
%         \mathcal{D}_t = \bigcup_{i=0}^{t} \mathcal{D}_i
%     \end{equation}

%     This ensures that newer models are trained on progressively larger datasets, mimicking an ML Ops retraining cycle.

%     \item \textbf{Intra-Chunk Time Scale (\( \tau \))}: Within each chunk \( t \), data points arrive sequentially, and ML Ops methods dynamically select a model from the available set \( \mathcal{M}_t = \{M_0, M_1, ..., M_t\} \) to process each incoming data point.

%     Formally, for each data point \( x_{\tau} \) in chunk \( t \):

%     \begin{equation}
%         M_{\tau} = \text{SelectModel}(\mathcal{M}_t, x_{\tau})
%     \end{equation}

%     where \( \text{SelectModel}(\cdot) \) represents the model selection rule determined by the chosen deployment strategy (e.g., A/B testing, MAB algorithms).
% \end{itemize}

\subsection{Evaluation of ML Ops Deployment Strategies}
Each of the deployment strategies described in the Background section was tested in this simulated environment:

\begin{itemize}
    \item \textbf{Naïve Model Replacement}: At each chunk boundary \( t \), the newly trained model \( M_t \) is immediately deployed for all intra-chunk decisions, ignoring prior models.
    \item \textbf{Validation-Based Selection}: The newly trained model \( M_t \) is deployed only if its validation score exceeds that of the currently deployed model.
    \item \textbf{A/B Testing}: \( M_t \) competes against \( M_{t-1} \) in a controlled experiment before deployment.
    \item \textbf{Epsilon-Greedy Multi-Armed Bandit}: The selection function \( \text{SelectModel}(\mathcal{M}_t, x_{\tau}) \) follows an \(\epsilon\)-greedy strategy, balancing exploration and exploitation.
    \item \textbf{Upper Confidence Bound (UCB) MAB}: Models are selected based on confidence intervals, prioritizing underexplored models.
    \item \textbf{Thompson Sampling MAB}: Model selection is performed probabilistically, with models chosen based on sampled performance estimates from Bayesian posteriors.
\end{itemize}

\subsection{Performance Metrics and Reward Functions}
We use different evaluation metrics tailored to the characteristics of each dataset:
\begin{itemize}
\item \textbf{Census dataset:} With a moderately imbalanced 70/30 class distribution, we use \textbf{Balanced Accuracy (BA)} as our primary evaluation metric. Balanced Accuracy is defined as the average of sensitivity (recall on positive class) and specificity (recall on negative class), effectively calculating $(TPR + TNR)/2$. This metric gives equal weight to both majority and minority classes regardless of their frequency in the dataset. For moderately imbalanced datasets like 70/30, BA provides a more informative assessment than regular accuracy, which can be misleadingly high when a model simply predicts the majority class. However, BA is not as extreme as PR-AUC, which is more appropriate for severe imbalances. This makes BA well-suited for the Census dataset, where we want to avoid majority class bias without completely discounting performance on the more common class.
\item \textbf{Fraud detection dataset:} With an extremely imbalanced 99/1 class distribution, we use \textbf{Area Under the Precision-Recall Curve (PR-AUC)} as our primary evaluation metric. PR-AUC plots precision against recall at various classification thresholds, focusing on the model's ability to correctly identify positive cases (fraud) while minimizing false positives. Unlike AUC-ROC, which can appear overly optimistic on highly imbalanced datasets due to its incorporation of true negatives, PR-AUC is particularly sensitive to performance on the minority class and doesn't reward the model for correctly identifying the abundant negative cases. This makes PR-AUC an ideal metric for fraud detection scenarios where correctly identifying the rare fraudulent transactions is of much higher importance than correctly classifying the numerous legitimate transactions.
\end{itemize}
For the reinforcement learning reward function in our UCB-MAB algorithm, we map these metrics to rewards to guide model selection. The choice of evaluation metric significantly impacts which models are favored during deployment, with PR-AUC encouraging models that better detect the rare fraud cases at the expense of potentially more false positives.

\subsubsection{Reward Function for Reinforcement Learning Methods}
For reinforcement learning-based methods, model selection is guided by a \textbf{reward function} that evaluates the effectiveness of the deployed model based on its observed performance. Instead of using the scoring function directly as the reward, we define the reward function based on whether a model's performance \textbf{improves} over time.

At each decision step, the reward is assigned as:

\begin{equation}
    R(M_{\tau}) =
    \begin{cases}
        1, & \text{if } f(M_{\tau}, \mathcal{D}_{\tau}) > f(M_{\tau-1}, \mathcal{D}_{\tau-1}) \\
        0, & \text{otherwise}
    \end{cases}
\end{equation}

where:
\begin{itemize}
    \item \( f(M_{\tau}, \mathcal{D}_{\tau}) \) represents the \textbf{score} of the deployed model on the incoming batch of data (metric is dataset dependent, as discussed above).
    \item A reward of 1 is assigned if the score on recent data is higher than the score on the previous data.
    \item A reward of 0 is assigned otherwise, preventing unnecessary model swaps.
\end{itemize}

This binary reward structure encourages reinforcement learning agents to favor models that consistently improve or maintain high performance while discouraging the deployment of models that fail to demonstrate better performance.

\subsubsection{Adapting the Reward Function for Regression Tasks}
While metrics discussed above are well-defined metric for classification problems, determining an equivalent reward function for \textbf{regression tasks} can be more complex. In regression settings, possible reward formulations include:

\begin{itemize}
    \item \textbf{Mean Squared Error (MSE) Reduction}: Reward based on the improvement in prediction error.
    \begin{equation}
        R(M_{\tau}) = - \left( MSE(M_{\tau}, \mathcal{D}_{\tau}) - MSE(M_{\tau-1}, \mathcal{D}_{\tau-1}) \right)
    \end{equation}
    where lower MSE values indicate better model performance.

    \item \textbf{R-Squared (\( R^2 \)) Improvement}: Measures how well the model explains variance in the target variable.
    \begin{equation}
        R(M_{\tau}) = R^2(M_{\tau}, \mathcal{D}_{\tau}) - R^2(M_{\tau-1}, \mathcal{D}_{\tau-1})
    \end{equation}

    \item \textbf{Dynamic Thresholding}: Using domain-specific constraints to determine acceptable error margins.
    \begin{equation}
        R(M_{\tau}) =
        \begin{cases}
            1, & \text{if } \text{Error Reduction} > \text{Threshold} \\
            0, & \text{otherwise}
        \end{cases}
    \end{equation}
\end{itemize}

Regression errors require careful selection of an appropriate reward function that aligns with the problem domain.

Future work could explore reinforcement learning methods for regression-based ML Ops by using adaptive reward shaping or uncertainty estimation techniques to better evaluate model improvements dynamically.

\section{Experimental Results and Discussion}
We evaluate the performance of different ML Ops deployment strategies on two large-scale classification datasets: the \textbf{Census Bureau USA dataset} \cite{census2023} and the \textbf{Bank Account Fraud dataset} \cite{fraud2022}. These datasets allow us to test how various deployment methods adapt to changing data distributions and assess whether reinforcement learning-based approaches outperform traditional static selection strategies.

\subsection{Evaluation Methodology}
As described in the experimental setup, each dataset was split into 8 sequential chunks to simulate a dynamic ML Ops environment. However, only \textbf{chunks 2-7} were used for evaluation, for the following reasons:

\begin{itemize}
    \item \textbf{Chunk 0}: No previously trained model exists to evaluate, making it unsuitable for model selection.
    \item \textbf{Chunk 1}: Only one model (\( M_0 \)) is available at this stage, meaning no selection decision is required.
\end{itemize}

Thus, all reported scores reflect model deployment performance across \textbf{chunks 2-7}.

\subsection{Performance Comparison on the Census Dataset}
The Census dataset was used to test model deployment performance on structured demographic and income-related classification tasks; the task is to classify earning amounts into buckets, creating a binary classification problem. The balanced-classification scores for each method are reported below:

\begin{table}[h]
    \centering
    \caption{Performance of Deployment Strategies on Census Dataset (BA Metric)}
    \begin{tabular}{|l|c|c|}
        \hline
        \textbf{Method} & \textbf{Overall Score} & \textbf{Chunk-Wise BA Scores} \\
        \hline
        Validation-Based Selection & 0.5995 & [0.6259, 0.6130, 0.6792, 0.5967, 0.5281, 0.5567] \\
        Naïve (Always Deploy New) & 0.5915 & [0.6259, 0.6130, 0.6305, 0.5967, 0.5281, 0.5567] \\
        A/B Testing & \textbf{0.6571} & [0.7070, 0.6790, 0.6964, 0.6535, 0.6001, 0.6092] \\
        Epsilon-Greedy ($\epsilon = 0.1$) & \textbf{0.6417} & [0.6400, 0.6509, 0.6630, 0.6301, 0.6300, 0.6365] \\
        UCB (\( c=0.1, b=300 \)) & 0.6398 & [0.6259, 0.6509, 0.6630, 0.6325, 0.6300, 0.6365] \\
        Thompson Sampling & 0.6117 & [0.6259, 0.6382, 0.6409, 0.6135, 0.5803, 0.5724] \\
        \hline
    \end{tabular}
    \label{tab: 1}
\end{table}

\subsubsection{Observations on Census Dataset Performance}
\begin{itemize}
    \item A/B Testing achieved the highest overall performance (0.6571), suggesting that statistical significance testing helps prevent model degradation.
    \item Thompson Sampling underperformed compared to other reinforcement learning methods, possibly due to excessive exploration causing performance fluctuations or the smaller size of the Census dataset now allowing enough time for the more complex RL method to fully optimize. We perform an in-depth breakout of Thompson sampling selection methods in the Fraud dataset.
    \item Epsilon-Greedy and UCB performed competitively, maintaining stable scores across chunks.
    \item The Naïve method (always deploying new models) had the lowest stability, suggesting that blindly deploying each new model can introduce instability.
\end{itemize}

\subsubsection{Epsilon-Greedy Model Selection in the Census Dataset}

To understand the impact of different exploration rates (\(\epsilon\)) on model selection and deployment performance, we analyze the behavior of \(\epsilon\)-greedy across different values of \(\epsilon\) in the Census dataset.

\begin{figure}[h]
    \centering
    \includegraphics[width=0.8\textwidth]{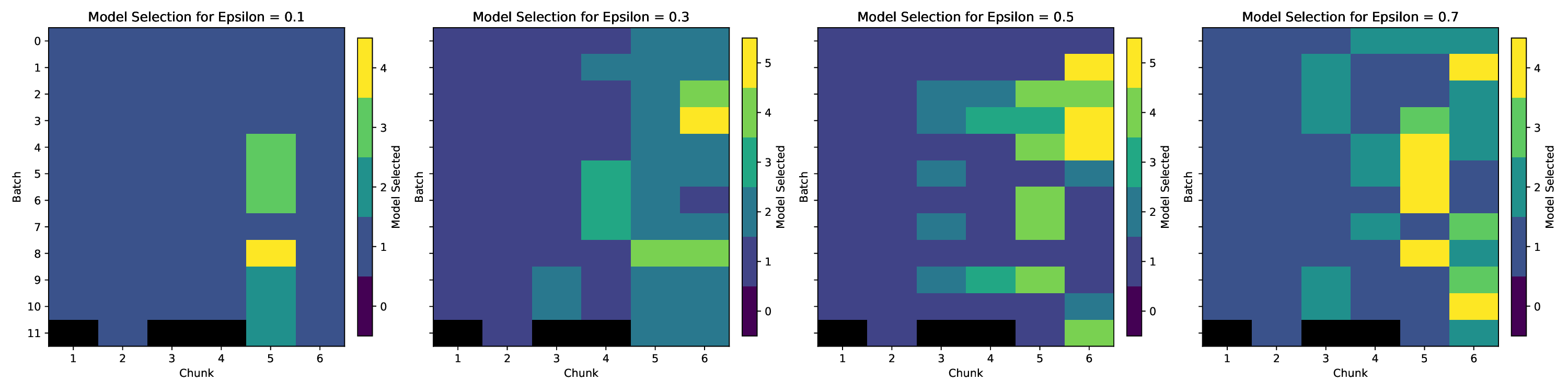}
    \caption{Model selection behavior across different values of \(\epsilon\) in the Census dataset.}
    \label{fig:epsilon_selection}
\end{figure}

\begin{figure}[h]
    \centering
    \includegraphics[width=0.8\textwidth]{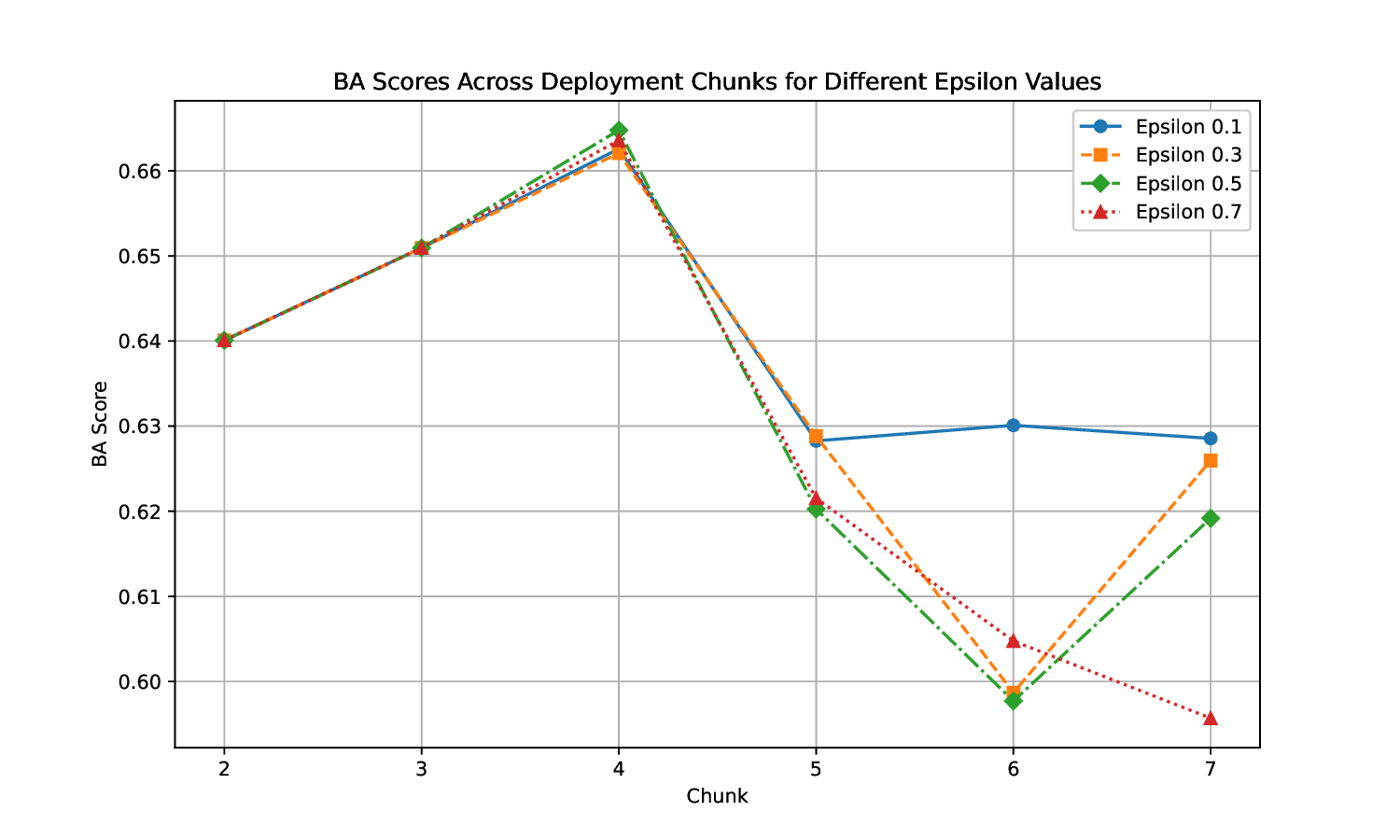}
    \caption{Balanced Classification scores across deployment chunks for different \(\epsilon\) values in the Census dataset.}
    \label{fig:auc_epsilon}
\end{figure}

\subsubsection{Analysis of Epsilon-Greedy in Census Deployment}
Figure~\ref{fig:epsilon_selection} shows how different values of \(\epsilon\) influence model selection across deployment chunks and batches.

\begin{itemize}
    \item At \(\epsilon = 0.1\), we see very little model switching, with the same model dominating most of the deployment.
    \item Increasing to \(\epsilon = 0.3\) and \(\epsilon = 0.5\) introduces more diversity in model selection, allowing newer models to be tested more frequently.
    \item At \(\epsilon = 0.7\), there is significant model switching, which could lead to instability if newer models do not generalize well.
\end{itemize}

Figure~\ref{fig:auc_epsilon} compares the BA performance across deployment chunks. 
\begin{itemize}
    \item Lower values of \(\epsilon\) (0.1, 0.3) result in more higher average BA scores, suggesting some of the earlier models were found to be stronger performers.
    \item Higher values (\(\epsilon = 0.5, 0.7\)) introduce more exploration but lead to more variance, with BA scores dropping sharply in later chunks.
    \item This suggests that \textbf{a moderate \(\epsilon\) provides a balance between stability and adaptability}, as it allows exploration without excessive performance drops.
\end{itemize}
These findings highlight the importance of tuning \(\epsilon\) appropriately based on dataset characteristics. For the Census dataset, \(\epsilon = 0.1-0.3\) appears to be a long term optimal trade-off, allowing sufficient adaptation while avoiding unnecessary model switching.

\subsection{Performance Comparison on the Fraud Dataset}
The Fraud dataset served as our testbed for evaluating model deployment strategies under challenging conditions: highly imbalanced classes (99/1 distribution), dynamically changing data distributions, and deliberately overfit models. In this environment, only earlier models demonstrated good generalization to production data, while later models exhibited poor performance. This scenario provided an ideal testing ground for assessing the adaptability of reinforcement learning-based selection methods, as it required them to quickly identify and favor the well-performing models amid several underperforming alternatives. Below, we report the PR-AUC scores for each method, which, as discussed earlier, is particularly well-suited for evaluating performance on highly imbalanced datasets where correctly identifying the rare positive class (fraud) is the primary concern:

\begin{table}[h]
    \centering
    \caption{Performance of Deployment Strategies on Fraud Dataset (PR-AUC)}
    \begin{tabular}{|l|c|c|}
        \hline
        \textbf{Method} & \textbf{Overall PR-AUC} & \textbf{Chunk-Wise PR-AUC Scores} \\
        \hline
        Epsilon-Greedy ($\epsilon = 0.3$, b = 700) & \textbf{0.0690} & [0.0622, 0.0530, 0.0961, 0.0477, 0.0858, 0.0862]  \\
        Validation-Based Selection & \textbf{0.0649} & [0.0572, 0.0564, 0.0815, 0.0607, 0.0560, 0.0931]  \\
        Thompson Sampling & 0.0548 & [0.0586, 0.0484, 0.0728, 0.0494, 0.0639, 0.0505] \\
        A/B Testing & 0.0545 & [0.0582, 0.0506, 0.0665, 0.0547, 0.0631, 0.0480] \\
        Naïve (Always Deploy New) & 0.0523 & [0.0612, 0.0515, 0.0668, 0.0583, 0.0512, 0.0554]  \\
        UCB (\( c=1, b=10000 \)) & 0.0529 & [0.0613, 0.0527, 0.0672, 0.0571, 0.0549, 0.0543]  \\
        \hline
    \end{tabular}
    \label{tab: 2}
\end{table}

\subsubsection{Observations on Fraud Dataset Performance}
\begin{figure}[H]
    \centering
    \includegraphics[width=0.8\textwidth]{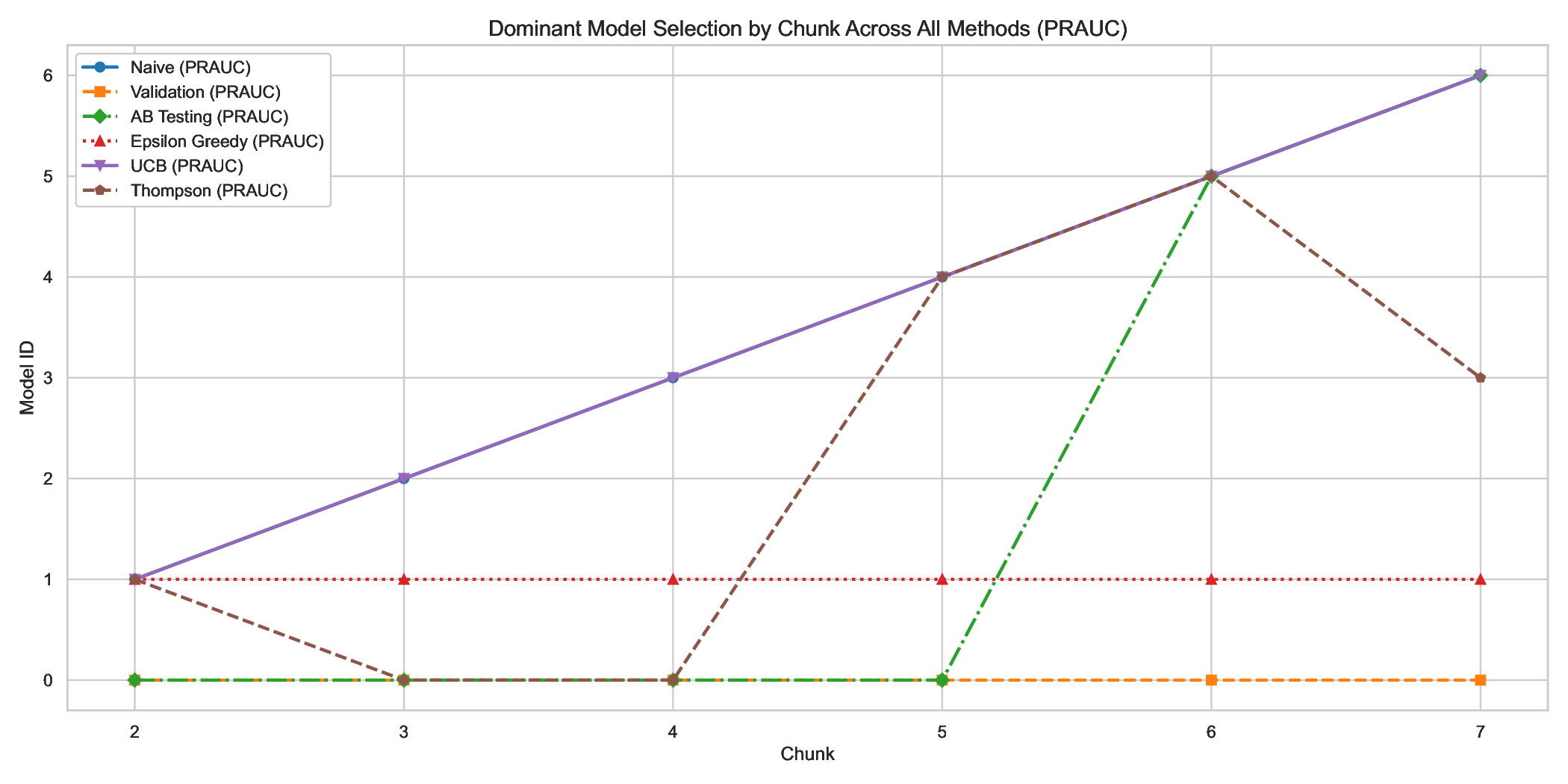}
    \caption{Dominant model selection by chunk across all methods (PR-AUC). The visualization reveals distinct selection patterns: Validation-Based consistently selects model 0 (orange dashed line), Epsilon-Greedy consistently selects model 1 (red dotted line), while UCB (purple solid line) and Thompson Sampling (brown dashed line) progressively explore higher-numbered models in later chunks. Note that the RL methods are showing majority model selection per chunk because they were able to switch models dynamically within a chunk.}
    \label{fig:modelselection}
\end{figure}

\begin{itemize}
    \item Epsilon-Greedy achieved the highest overall PR-AUC (0.0690), demonstrating its effectiveness at balancing exploration and exploitation in the challenging fraud detection context. The epsilon greedy method was able to consistently select model 1, the best performing model, throughout the deployment process as shown in Figure \ref{fig:modelselection}. This strategic consistency was key in maximizing the performance score across the production deployment simulation. The method's controlled exploration rate ($\epsilon = 0.3$, note, slightly higher exploration value than Census dataset) provided sufficient safety against concept drift while maintaining exploitation of the best-performing model.
    \item Validation-Based Selection performed remarkably well (0.0649 PR-AUC), showing that careful evaluation on validation data can lead to effective model selection. This method was particularly strong in the later chunks (3 and 6), suggesting good adaptability to evolving fraud patterns. As evident in Figure \ref{fig:modelselection}, Validation exclusively selected model 0 throughout the deployment process. While model 0 was a strong performer, the method never switched to model 1 (the top performer identified by Epsilon-Greedy), which explains its slightly lower overall performance. This highlights a potential limitation of validation-based approaches when validation data doesn't perfectly reflect production data characteristics.
    \item Thompson Sampling and A/B Testing achieved moderate performance (0.0563 and 0.0545 PR-AUC respectively), offering better results than the naive approach but falling short of the top methods. Their similar chunk-wise patterns suggest comparable exploration strategies. However, Figure \ref{fig:modelselection} reveals a fascinating progression in Thompson's model selection behavior: it favors model 0 in early chunks but progressively shifts toward higher-numbered models (3, 4, and eventually 5) in later chunks. This dynamic adaptation pattern demonstrates Thompson's continuous learning and willingness to explore new models as data distributions evolve, though this exploration comes at the cost of immediate performance.
    \item The Naïve (Always Deploy New) approach performed poorly (0.0523 PR-AUC), confirming that blindly deploying new models without performance considerations is suboptimal in fraud detection scenarios where class distributions are highly imbalanced. The approach follows a predefined pattern visible in Figure \ref{fig:modelselection}, where it consistently deploys the newest model (model n for chunk n+1). This strategy fails to recognize that newer models aren't necessarily better, especially in domains with concept drift and class imbalance where specialized models may perform better across multiple time periods.
    \item UCB was provided a higher c value than in the Census dataset to understand how the exploration settings play a role in its performance. Figure \ref{fig:modelselection} reveals UCB's distinctive linear progression through models—starting with model 1 and methodically increasing to models 2, 3, 4, 5, and finally 6. This suggests UCB's confidence bounds led it to systematically explore newer models in sequence, failing to recognize model 1's superior performance. The exploration parameter (c = 1) prioritized exploration over exploitation, causing it to disregard the top-performing model in favor of trying newer alternatives. In this situation, UCB approximates a naive selection method. This demonstrates, generally, that the RL based selection methods are flexible enough to approximate any of the standard selection techniques. 
    \item Across all methods, Figure \ref{fig:modelselection} reveals striking diversity in model selection strategies. The flat lines (Validation and Epsilon-Greedy) represent consistent selection of a single model, while ascending patterns (UCB, Thompson, and AB Testing) show progressive exploration of newer models. This visualization effectively captures the fundamental exploration-exploitation tradeoff in reinforcement learning approaches to model deployment.
\end{itemize}

\subsubsection{Thompson Sampling Model Selection in the Fraud Dataset}

Thompson Sampling demonstrates a unique model selection pattern that reveals important insights about exploration-exploitation dynamics in reinforcement learning systems. While not achieving the highest overall PR-AUC, it provides valuable insights into adaptive model selection.

\begin{figure}[h]
    \centering
    \includegraphics[width=0.8\textwidth]{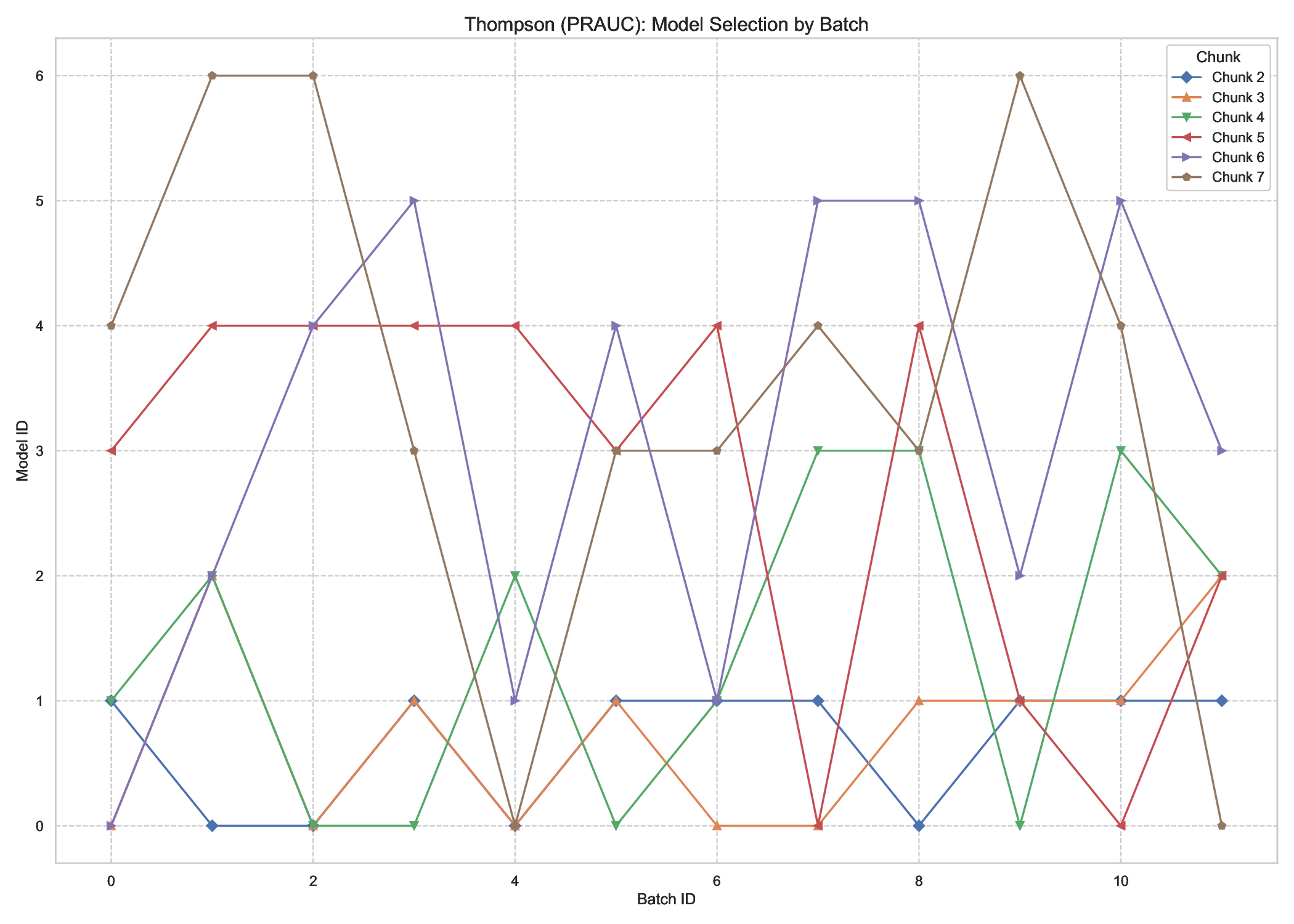}
    \caption{Thompson Sampling's batch-level model selection across all chunks, showing the dynamic switching pattern between models during the learning process. Each line represents a dataset chunk, and each point on the line a unique batch at which the model was able to decide to swap models.}
    \label{fig:thombybatch}
\end{figure}

Figure \ref{fig:thombybatch} illustrates Thompson Sampling's batch-level model selections, revealing several interesting behaviors. Unlike other methods that quickly converge to a single model, Thompson Sampling maintains exploration throughout the deployment period, regularly testing multiple models even in later batches. This behavior is particularly evident in Chunk 6, where it consistently alternates between model 5 and various alternatives, and in Chunk 7, where it explores higher-numbered models (6) while occasionally reverting to models 1, 3, and 4. This willingness to explore explains why Thompson achieved moderate but not top performance, but also shows how an RL method can be tuned to an ML Ops environment in which managers may prefer to lean towards deploying newer models while still having a dynamic safety net. 

As shown in Figure \ref{fig:modelselection}, Thompson Sampling (brown dashed line) initially selects Model 1 and 0 for early chunks, similar to Epsilon-Greedy and Ab Testing, but then shifts toward higher-numbered models in later chunks. This stands in contrast to Validation-Based Selection (orange dashed line), which consistently selects Model 0, and Epsilon-Greedy (red dotted line), which maintains Model 1 throughout. This suggests Thompson's posterior probability updates allowed it to detect changing data distributions that favored newer models in later chunks.

\begin{figure}[h]
    \centering
    \includegraphics[width=0.8\textwidth]{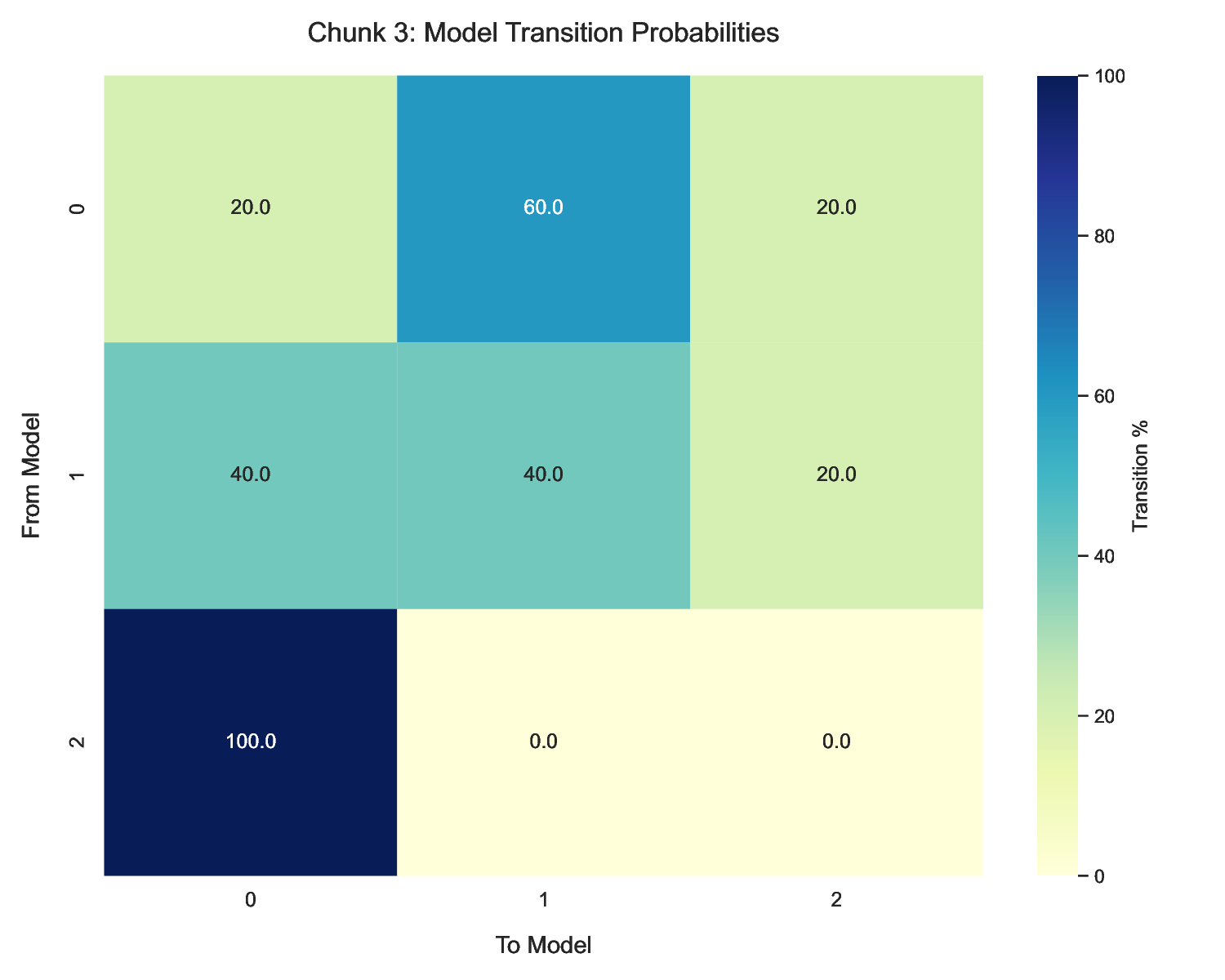}
    \caption{Transition probabilities between selected models in Chunk 3 using Thompson Sampling. Rows represent the model selected at time t and columns represent the model selected at time t+1. Each cell shows the percentage of transitions observed between model pairs. For example, after selecting Model 2, the algorithm transitioned to Model 0 in 100\% of cases, indicating strong instability in Model 2 selections. In contrast, Model 1 exhibited higher persistence, remaining selected in 40\% of transitions.}
    \label{fig:chunk3probs}
\end{figure}

The transition probability matrices provide deeper insights into Thompson Sampling’s decision-making process. In Chunk 3 (Figure 5), Thompson demonstrates dynamic switching behavior across Models 0, 1, and 2. While Model 1 exhibits moderate stability—being reselected 40\% of the time—it also frequently transitions to other models, indicating balanced exploration. Notably, Model 2 is never selected in consecutive batches and always transitions to Model 0, highlighting its perceived underperformance. This early-stage exploratory behavior enables Thompson Sampling to refine its posterior estimates by systematically sampling across the available model space.

\begin{figure}[h]
    \centering
    \includegraphics[width=0.8\textwidth]{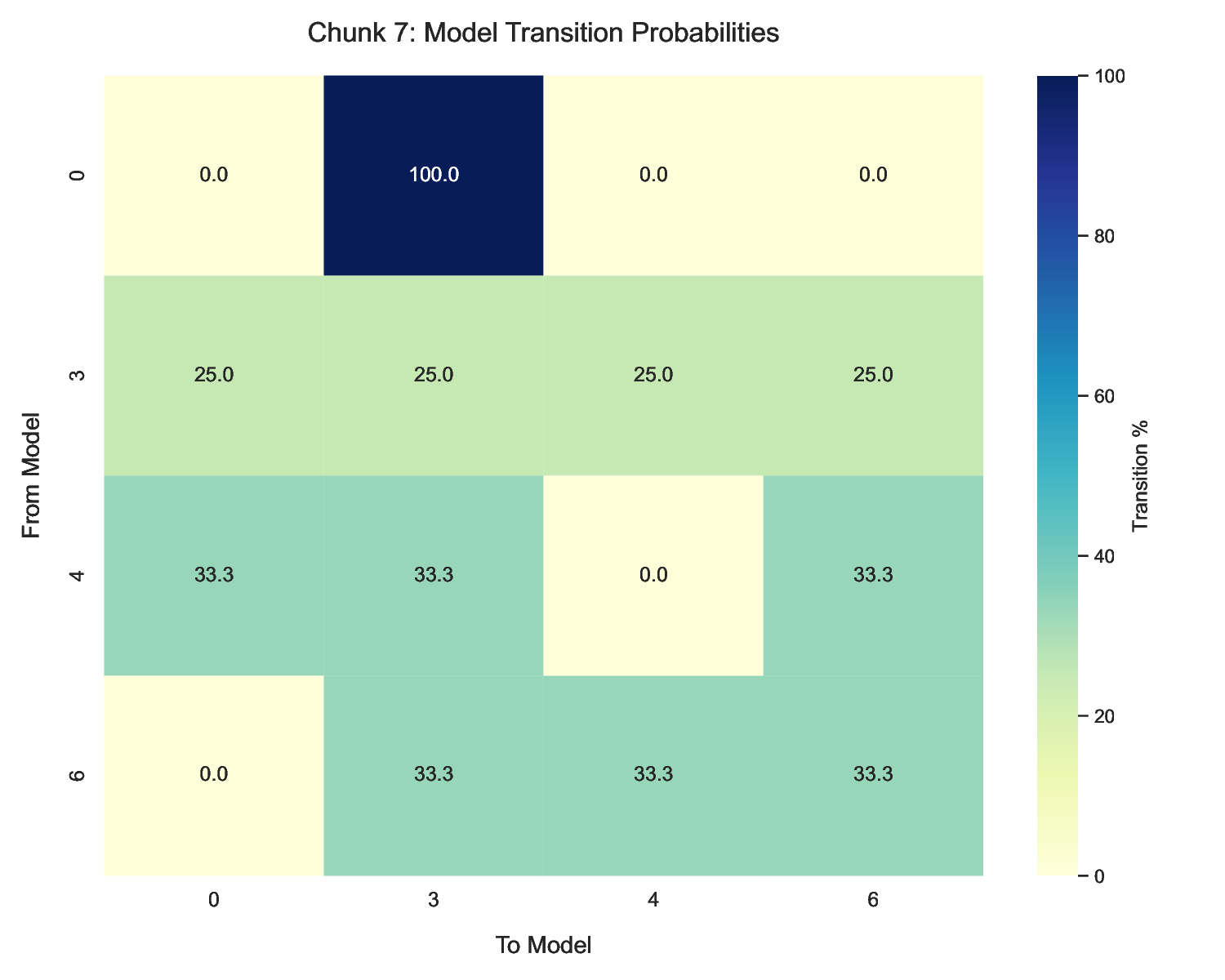}
    \caption{Model transition probabilities for Chunk 7 under Thompson Sampling. Each cell represents the percentage of times the algorithm transitioned from the model on the Y-axis to the model on the X-axis. Model 3 appears frequently and transitions evenly to all available models, reflecting high exploration. Model 4 shows a similar trend but never loops back to itself. Notably, Model 0 is deterministic—always followed by Model 3—while Model 6 transitions are spread evenly across Models 3, 4, and 6, suggesting a lack of clear model dominance at this stage.}
    \label{fig:chunk7probs}
\end{figure}

By Chunk 7 (Figure \ref{fig:chunk7probs}), Thompson Sampling’s transition matrix reveals a more exploratory and less deterministic pattern across models 0, 3, 4, and 6. Model 0 exhibits a fully deterministic transition, moving to Model 3 in 100\% of cases, suggesting a lack of confidence in Model 0’s sustained performance, notable considering this is one of the stronger performing models overall. In contrast, Model 3 transitions are evenly distributed across all available options (25\% each), indicating active exploration. Model 4 alternates between Models 0, 3, and 6 with equal probability (33.3\%), while never transitioning to itself. Similarly, Model 6 distributes its selections evenly across Models 3, 4, and 6. These transition dynamics suggest that Thompson has not yet converged on a dominant model in Chunk 7, instead maintaining a balanced strategy that continues to explore multiple viable options.

\textbf{Key Takeaways:}
\begin{itemize}
    \item Thompson Sampling maintains exploration throughout deployment, unlike methods that quickly converge to a single model. This makes it more robust to concept drift but potentially sacrifices short-term performance.
    
    \item The transition matrices revealed in Figures \ref{fig:chunk3probs} and \ref{fig:chunk7probs} show that Thompson develops increasingly confident model preferences in later chunks.
    
    \item Unlike Epsilon-Greedy, which achieved the highest PR-AUC by consistently selecting Model 1 as seen in Figure \ref{fig:modelselection}, Thompson progressively shifted to newer models. This suggests the data distribution may have been changing in ways that Thompson detected but that didn't immediately translate to PR-AUC improvements.
    
    \item Thompson's willingness to explore Model 5 and Model 6 in later chunks (which no other method selected as dominant) demonstrates its value for discovering potentially valuable models that more exploitative approaches might miss, as evidenced in Figure \ref{fig:thombybatch}.
    
    \item The moderate overall performance of Thompson (0.0548 PR-AUC) compared to Epsilon-Greedy (0.0690) highlights the classic exploration-exploitation tradeoff: Thompson sacrificed some immediate performance for potentially better long-term adaptation.
\end{itemize}

This analysis suggests that while Thompson Sampling didn't achieve the highest PR-AUC in this specific fraud detection scenario, its dynamic selection behavior provides valuable insights for deploying ML models in environments where data distributions evolve over time. Its ability to balance exploration with increasingly focused exploitation makes it a compelling choice for scenarios where robustness to distribution shifts is more important than maximizing short-term performance metrics.

\subsection{Overall Findings and Takeaways}
Our results highlight key differences in model deployment performance across datasets and demonstrate the advantages of reinforcement learning-based ML Ops strategies. Importantly, our goal was not to train the best possible predictive model but rather to optimize the deployment process. As such, models were trained using standard techniques, including one-hot encoding, feature scaling, and XGBoost classifiers, to focus on evaluating the effectiveness of different deployment strategies. In both datasets, the top two performing methods were highlighted (\ref{tab: 1}, \ref{tab: 2}); epsilon greedy selection appeared in both of the datasets as a top performing method, whereas, none of the traditional selection metrics appeared consistently across both datasets as top performers.

% \subsubsection{UCB Discussion}
% We conduct an analysis of the UCB selection mechanism on the Fraud dataset, focusing on how the exploration parameter \( c \) influences model selection behavior over time. Figure~\ref{fig:ucb_heatmaps} visualizes the proportion of model selections across chunk iterations for three values of \( c \in \{0.1, 0.5, 1.0\} \).

% As expected, lower values of \( c \) strongly emphasize exploitation. At \( c = 0.1 \), UCB deterministically selects Model 1 for every chunk, mirroring greedy behavior with no exploration, similar to the top performing epsilon greedy method in the fraud dataset. As \( c \) increases, exploration emerges: at \( c = 0.5 \), the algorithm begins considering alternative models, favoring Model 2 and Model 3 in later chunks. However, the most striking behavior occurs at \( c = 1.0 \), where model selection begins to always select the newest model. This suggests UCB approximates a naïve or uniform selection policy at high \( c \) values, and interestingly, that the RL methods can approximate any of the standard selection methods provided the correct level of exploration is selected.

% \begin{figure}[h]
%     \centering
%     \includegraphics[width=0.8\textwidth]{fraud_ucb.eps}
%     \caption{UCB Model selection behavior across different values of \textit{c} in the Fraud dataset. As was seen in \ref{fig:modelselection}, UCB approximates a naive method at a certain point of exploration.}
%     \label{fig:ucb_heatmaps}
% \end{figure}

\subsubsection{Key Takeaways}
\begin{itemize}
    \item The Census dataset is smaller than the Fraud dataset, which affects deployment strategy performance. However, in both datasets, an appropriate choice of exploration parameter allowed the RL methods to outperform the traditional ML Ops methodologies. 
    \item Epsilon-Greedy performed best overall, matching or exceeding the top-performing method in both datasets. 
    \item Reinforcement learning-based strategies (Epsilon-Greedy, UCB, Thompson Sampling) provided adaptive deployment strategies, particularly in environments where models exhibited overfitting (fraud detection).
    \item Static heuristics like validation-based selection struggled with generalization issues—they frequently deployed overfitted models in fraud detection.
    \item A/B Testing performed well in stable environments (Census dataset) but struggled with high-variance, imbalanced data (fraud detection).
\end{itemize}

These findings suggest that reinforcement learning-based approaches offer a more robust solution for dynamic model deployment, especially in scenarios where model performance varies significantly over time.

\section{Conclusion}
In this study, we evaluated the effectiveness of reinforcement learning-based approaches for model deployment in ML Ops environments across two diverse datasets. Our findings demonstrate that RL methods, particularly Epsilon-Greedy and UCB algorithms, can match or exceed traditional deployment strategies in dynamic, real-world settings.

The observed performance patterns suggest several important insights for ML practitioners. First, RL approaches show superior adaptability in environments with model drift and overfitting issues. This was particularly evident in the Fraud dataset, where traditional approaches struggled with the gap between validation and production performance. Second, the efficacy of deployment strategies is heavily influenced by dataset characteristics—with RL methods demonstrating greater advantages in larger, more complex datasets where exploration-exploitation trade-offs become more critical.

Our detailed analysis of exploration parameters (such as $\epsilon$ in Epsilon-Greedy and $c$ in UCB) revealed that these hyperparameters require thoughtful tuning based on both dataset size and model quality. The optimal exploration rate varies depending on the stability of the underlying data distribution and the relative performance differences between candidate models.

From a practical standpoint, our research suggests that ML Ops pipelines could benefit significantly from incorporating RL-based deployment strategies, especially in domains prone to drift. These methods offer a more autonomous approach to model selection, reducing the need for manual intervention and potentially shortening the feedback loop between model training and deployment.

Notably, our work investigates a relatively unexplored environment in which production ML systems can face IID-like time series problems—where data arrives sequentially but may follow patterns that aren't strictly time-dependent. This hybrid environment, positioned between classical time series and traditional IID settings, remains underexplored from a quantitative standpoint. The reinforcement learning approaches we've evaluated provide an initial framework for addressing this unique challenge, but numerous unexplored directions remain. These include developing specialized reward functions that account for both immediate performance and temporal stability, creating adaptive exploration strategies that respond to detected distribution shifts, and designing more sophisticated credit assignment mechanisms for delayed performance signals.

Future research should explore how these approaches scale to larger model ecosystems with dozens or hundreds of candidate models, and how they perform in multi-objective optimization scenarios where business metrics extend beyond standard performance indicators. Additionally, investigating how RL strategies can be combined with drift detection mechanisms could further enhance ML Ops automation. Another promising direction would be exploring how these approaches handle heterogeneous model architectures, where the candidate models might differ significantly in their underlying approaches.

Overall, our work highlights the potential of reinforcement learning as a valuable addition to the ML Ops toolkit, providing a more adaptive and robust framework for model deployment in dynamic, real-world environments.

%Bibliography
\bibliographystyle{unsrt}  
\bibliography{references}

\end{document}